\documentclass{llncs}

\usepackage{makeidx}
\usepackage[acronym, shortcuts]{glossaries}
\newacronym{acronym:slam}{SLAM}{simultaneous localisation and mapping}
\newacronym{acronym:sfm}{SfM}{structure-from-motion}
\newacronym{acronym:mvs}{MVS}{multi-view stereo}
\newacronym{acronym:icp}{ICP}{iterative closest point}
\newacronym{acronym:gt}{GT}{ground truth}
\newacronym{acronym:tsdf}{TSDF}{truncated signed distance function}
\newacronym{acronym:rms}{RMS}{root mean square}
\newacronym{acronym:ate}{ATE}{absolute trajectory error}
\usepackage{graphicx}
\graphicspath{{images/}}
\usepackage[inline, shortlabels]{enumitem}
\usepackage{amssymb}
\usepackage{hyperref}
\hypersetup{
    colorlinks,
    citecolor=black,
    filecolor=black,
    linkcolor=black,
    urlcolor=black
}

\DeclareMathOperator{\dist}{dist} 

\begin{document}

\title{A comparative study of breast surface reconstruction for aesthetic outcome assessment}
\titlerunning{Breast surface reconstruction}

\author{Ren\'{e} M.\ Lacher\inst{1} 
\and Francisco Vasconcelos\inst{1}
\and David C.\ Bishop\inst{2}
\and Norman R.\ Williams\inst{3}
\and Mohammed Keshtgar\inst{4}
\and David J.\ Hawkes\inst{1}
\and John H.\ Hipwell\inst{1}
\and Danail Stoyanov\inst{1}}


\authorrunning{Ren\'{e} Lacher et al.}

\tocauthor{Ren\'{e} Lacher, Francisco Vasconcelos, David Bishop, Norman Williams, Mohammed Keshtgar, David Hawkes, John Hipwell, Danail Stoyanov}

\institute{Centre for Medical Image Computing, University College London, UK\and
Medical Photography, Royal Free \& University College Medical School, London, UK\and
Surgical \& Interventional Trials Unit, University College London, UK\and
Royal Free London Foundation Trust, London, UK\\
\email{\{rene.lacher.13, f.vasconcelos, david.bishop, norman.williams, m.keshtgar, d.hawkes, j.hipwell, danail.stoyanov\}@ucl.ac.uk}}

\maketitle

\begin{abstract}
Breast cancer is the most prevalent cancer type in women, and while its survival rate is generally high the aesthetic outcome is an increasingly important factor when evaluating different treatment alternatives. 3D scanning and reconstruction techniques offer a flexible tool for building detailed and accurate 3D breast models that can be used both pre-operatively for surgical planning and post-operatively for aesthetic evaluation. This paper aims at comparing the accuracy of low-cost 3D scanning technologies with the significantly more expensive state-of-the-art 3D commercial scanners in the context of breast 3D reconstruction. We present results from 28 synthetic and clinical RGBD sequences, including 12 unique patients and an anthropomorphic phantom demonstrating the applicability of low-cost RGBD sensors to real clinical cases. Body deformation and homogeneous skin texture pose challenges to the studied reconstruction systems. Although these should be addressed appropriately if higher model quality is warranted, we observe that low-cost sensors are able to obtain valuable reconstructions comparable to the state-of-the-art within an error margin of 3\,mm.

\end{abstract}

\section{Introduction}

Breast cancer affects women worldwide and recent incidence figures report $1.8$ million new cases diagnosed per annum making breast cancer the most common cancer in females \cite{naghavi15global}. Roughly two thirds of patients choose a less invasive lumpectomy combined with radiotherapy over a complete breast removal. Breast conserving surgery achieves comparable oncological outcomes while preserving as much of the healthy breast tissue as possible yielding a superior cosmetic outcome. However, approximately $29\%$ of patients are left with a suboptimal - that is fair or poor - aesthetic result \cite{cardoso14assessing}. The increasingly favourable prognosis for a majority of patients and the link between aesthetic outcome and quality of life places a high importance on outcome assessment, planning and simulation to identify and correlate contributing factors. Nowadays, cosmetic outcome assessment still lacks standardisation in clinical practice and it is often undertaken as an expert evaluation of the patient in person or via 2D photography \cite{cardoso14assessing}. This process is costly, time-consuming and inherently subjective. Recent technological advances and the maturity of computer vision based 3D surface imaging technology allow high-fidelity 3D breast surface capture for aesthetic assessment. Nevertheless, commercial systems are typically infrastructure-heavy and expensive \cite{oconnell15review}.

\begin{figure}[ht]
\includegraphics[width=\columnwidth]{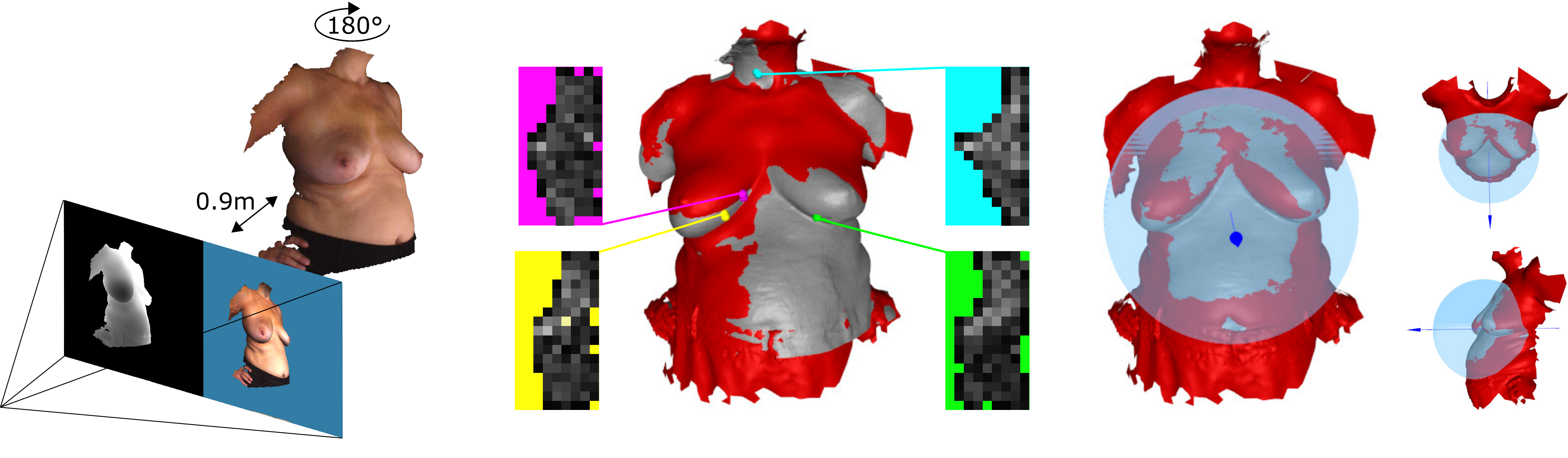}
\caption{Left: Illustration of synthetic data generation. GT model is placed in virtual scene and projected into simulated RGBD camera. A sequence of images is generated while the model spins around its y-axis. Middle, right: The two stages of the automatic registration for validation. Vertex-wise spin images are matched between the downsampled source and target model for coarse alignment. Point-to-plane ICP is limited to all source vertices inside a central sphere.}
\label{fig:synthetic_data_automatic_registration}
\end{figure}
The aim of this paper is to show low-cost breast reconstruction on a standard desktop computer utilising free software and mass-market camera technology. Four dense \gls{acronym:slam} and \gls{acronym:sfm} systems without any shape prior are compared. A three-fold experiment comprising synthesized and real phantom and patient data was conducted. Results are validated against submillitmeter-accurate gold standard models from commercial scanning systems, as well as, in the synthetic case, \gls{acronym:gt} camera trajectories.
Even though low-cost depth cameras have previously been used for reconstructing breast surfaces, existing methods manually select sparse keyframes rather than building a model from full frame-rate video \cite{wheat14development,costa14tessellation}.
The characteristics and differences between both generations of Kinect have been comprehensively explored by \cite{sarbolandi15kinect} but no study has been conducted assessing the sensors for breast surface reconstruction as other comparative studies did with respect to prototypical or commercial 3D scanners \cite{patete13comparative}.
Our results \begin{enumerate*}[(i)] \item indicate that low-cost systems produce promising results that can potentially be used clinically and \item significantly increase the use of objective measures for aesthetic planning and assessment \end{enumerate*}.

\section{Methods}

We compare 3D breast reconstruction using two high-precision scanning solutions, a structured-light handheld Artec Eva scanner for the phantom and a single shot 3dMD stereophotogrammetry system for patients, against reconstructions obtained from a low-cost RGBD Microsoft Kinect v1 and Kinect v2. Such systems rely upon proprietary software, whereas we only use freeware and open-source code. Kinect data sets are 3D reconstructed through three different state-of-the-art algorithms for RGBD data (ElasticFusion \cite{whelan15elasticfusion},  InfiniTAM \cite{prisacariu14framework},  Lacher et\ al.\ \cite{lacher15low}), along with an algorithm purely using RGB data (VisualSfM \cite{wu13towards,furukawa10accurate}).
ElasticFusion features a joint geometric and photometric tracking component and time-windowed non-rigid loop closure strategies fusing data into a dense surfel cloud. A surfel extends the notion of a point to a locally planar patch of some radius. InfiniTAM is an extensible \gls{acronym:slam} framework integrating a hierarchical \gls{acronym:tsdf} volume representation while sharing its core functionality with the works of \cite{newcombe11kinectfusion}. Lacher et\ al.\ introduce an explicit clipping of unreliable measurements and extend the latter system by a pose graph diffusion step. 
VisualSfM is an incremental SIFT-based \gls{acronym:sfm} alternating bundle adjustment and re-triangulation followed by quasi-dense \gls{acronym:mvs} expanding surface patches in their projective neighbourhoods through optimisation of photometric consistency subject to visibility constraints.
Given our aim of evaluating how low-cost technologies fare against high-accuracy commercial scanning systems, their reconstructions serve as \gls{acronym:gt}. The \gls{acronym:gt} is registered against reconstructions from aforementioned techniques which are using Kinect data. This registration employs a common \gls{acronym:icp} with point-to-plane error metric initialised by a robust matching of spin-images \cite{johnson97spin}. In the following, two 3D reconstruction errors are measured:
\begin{description}
\item[Surface-to-surface distance:] The smallest distance $\min_j \dist_{ptt}(\boldsymbol{p}^s_i, \boldsymbol{f}^t_j)$ of all N source points $\boldsymbol{P}^s = \{\boldsymbol{p}^s_i \in \mathbb{R}^3\}_{i=1}^N$ to the closest of the $M$ target mesh triangular faces $\boldsymbol{F}^t = \{\boldsymbol{f}^t_j \in [1..N]^3\}_{j=1}^M$. This distance is computed using the exact point-to-triangle algorithm proposed as the 2D method by \cite{jones953d} in a naively GPU-parallelised reimplementation.
\item[Surface normal deviation:] The normal error is defined as the difference in normal orientation $\cos^{-1}(\langle\boldsymbol{n}^s_i,\boldsymbol{\hat{n}}\rangle)$ between the normal of source vertex $\boldsymbol{p}^s_i$ and the bilinearly or barcycentrically interpolated normal $\boldsymbol{\hat{n}} \in \mathbb{R}^3$ at the intersection with the closest target triangle $\boldsymbol{f}^t_j$.
\end{description}

With both metrics, error computation is heuristically confined to a region of interest covering the breasts in a bounding sphere. This sphere is centred at the intersection of a line parallel to the z-axis passing through the target mesh's centroid as depicted in Fig.\,\ref{fig:synthetic_data_automatic_registration}. This way, parts with foreseeable large deformation are excluded and a direct comparison of methods is possible as errors are accumulated over an identical region. Likewise, source points matching to a boundary target triangle indicating a non-overlap region are excluded from error statistics. Boundary triangles are determined by finding all triangles with one or more single edges.

\begin{figure}[t!]
\centering
\includegraphics[width=\columnwidth]{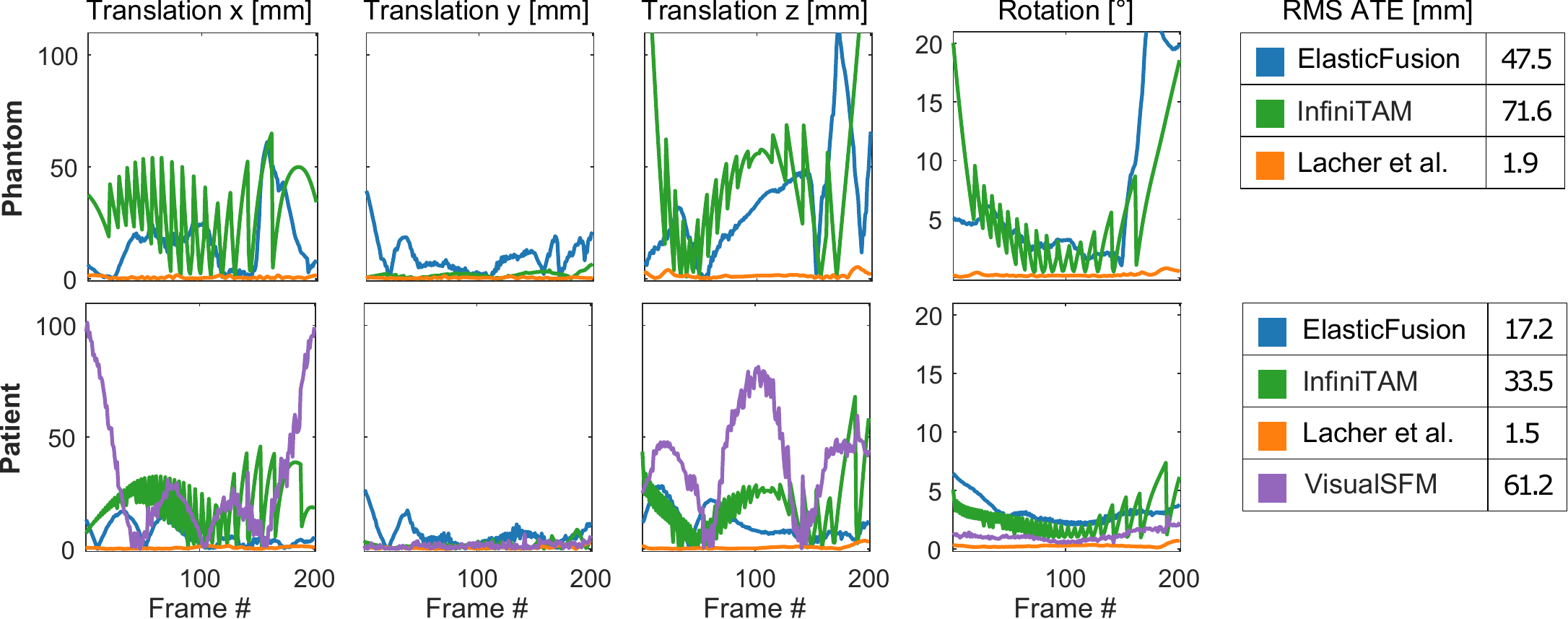}
\caption{Pose error profile in synthetic data experiment. The figure shows the quantified translational and rotational error plotted against estimated camera poses.}
\label{fig:synthetic_pose_error}
\end{figure}

Since all methods also estimate the sensor motion trajectories for each acquisition, translational and rotational camera pose errors are reported for reconstructions from synthetic RGBD for which \gls{acronym:gt} trajectory data is available (see Fig.\,\ref{fig:synthetic_pose_error}). The rotational error is extracted as the shortest arc angle $2\cos^{-1}(\hat{q}_w)$ of an interpolating quaternion $\hat{q} = q^{s} \cdot \bar{q}^{t}$ between two corresponding camera orientations in normalized quaternion form $q^{s}$ and $q^{t}$, where $\bar{q}$ is denoting the quaternion conjugate. For an overall trajectory error score we evaluate the \gls{acronym:rms} \gls{acronym:ate} as proposed by \cite{sturm12benchmark}.

\section{Experiments and Results}

Before acquisition, Kinect intrinsic and extrinsic geometric parameters were calibrated using a checkerboard grid. Albeit, both Kinects provide mid-resolution depth, however the Kinect v2 uses a third fewer pixels and is equipped with shorter lenses, yet streams RGB in full HD. Compliant with protocol, subjects are positioned in front of a blue homogeneous background at 0.9\,m distance from a static tripod-mounted Kinect. Phantom data was recorded by placing an anthropomorphic mannequin on an electric rotation platform. Patients are asked to stand hands on hips and slowly self-rotate on the spot while the $180^\circ$ RGBD sequence is recorded. The region around the patient's breast is assumed to remain rigid for the duration of the acquisition, which allows the fusion of all frames into a single 3D breast model. 24 data sequences from 12 patients with a mean scanning duration of 11.5$\pm$2.9\,s (608$\pm$171\,frames) were selected from a larger cohort of patients undergoing breast conservative surgery including patients of varying cup sizes and age groups. This selection was made such that no two adjacent timestamps in any RGBD sequence exceeded 100\,ms.
Additionally, synthetic phantom and patient data was created by placing the respective \gls{acronym:gt} model in a virtual scene with two point light sources (see Fig.\,\ref{fig:synthetic_data_automatic_registration}). RGBD images were rendered into a simulated camera utilising customised framebuffer and shader objects for depth and normal computation in camera coordinates resulting in a half circular in-plane camera trajectory in compliance with the clinical acquisition protocol. Synthetic data sequences only undergo rigid motion and are free of noise and lens distortions. The synthetic phantom was artificially textured with a uniform skin tone.

\begin{figure}[t]
\centering
\includegraphics[width=\columnwidth]{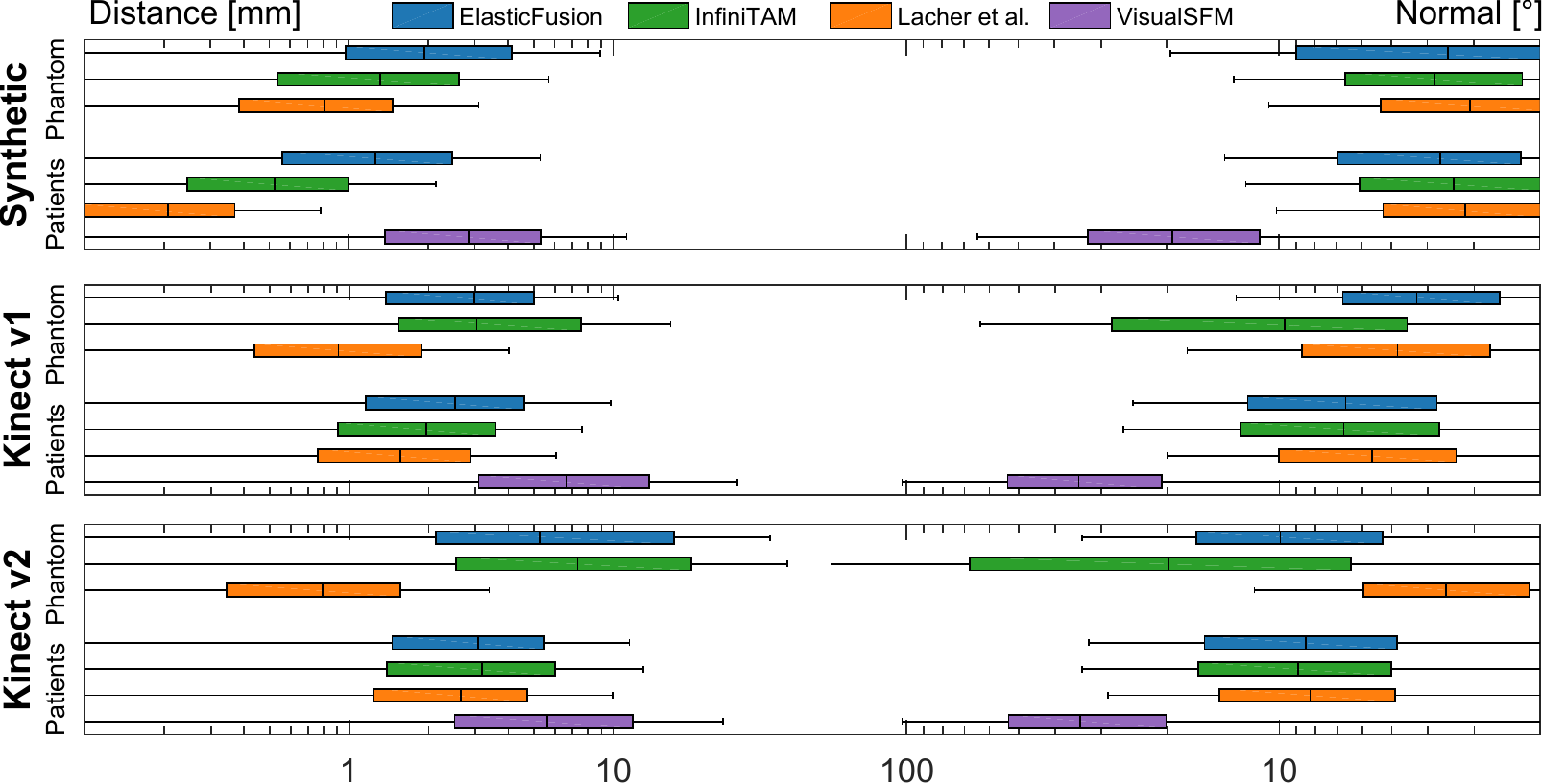}
\caption{Surface-to-surface error distributions horizontally grouped into syntethic, Kinect v1 and v2 results. The bars on the left correspond to mean surface distances, the bars on the right to mean surface normal deviation.}
\label{fig:surface_error_boxplot}
\end{figure}
\begin{figure}[t]
\centering
\includegraphics[width=\columnwidth]{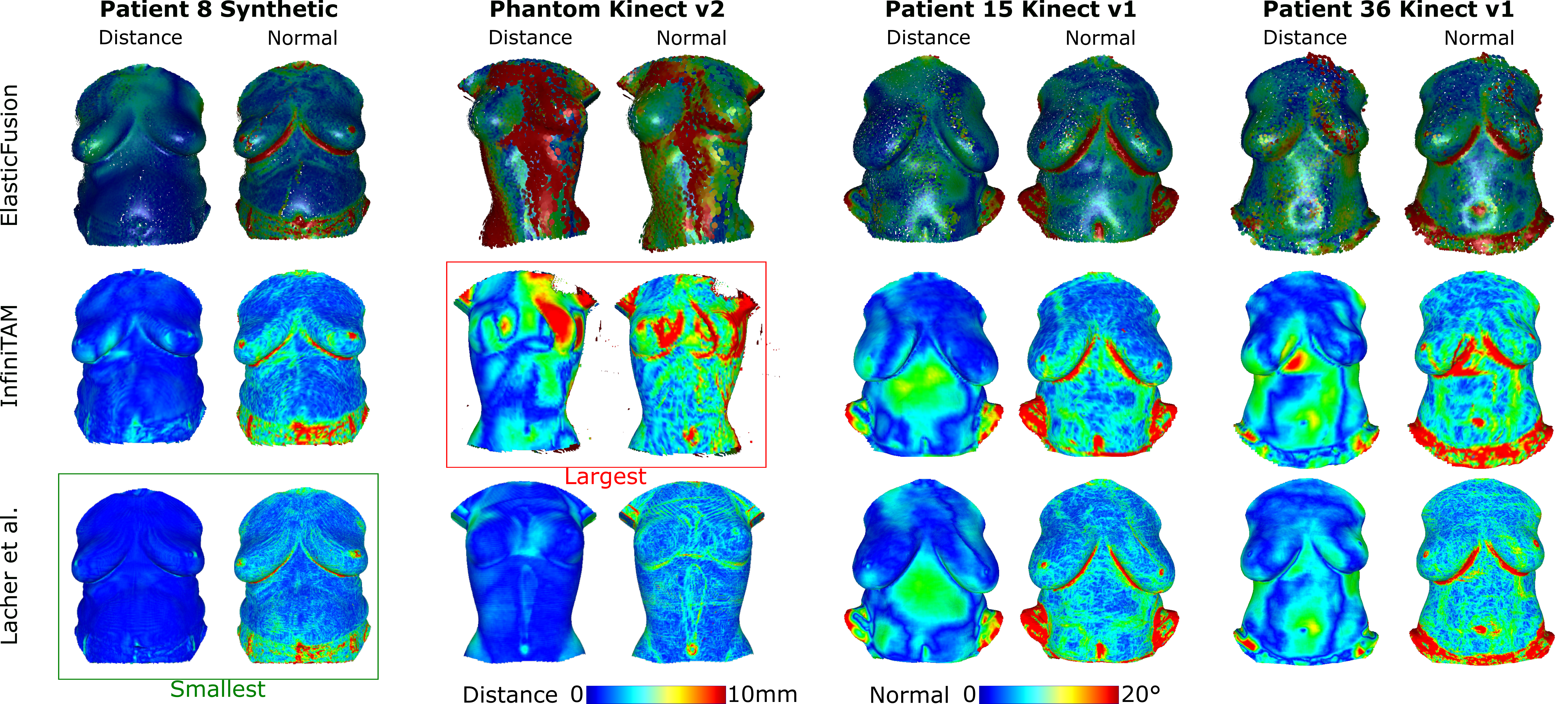}
\caption{Colourmapped surface-to-surface distance to the left of surface normal deviation for all methods excluding VisualSfM. We display the data sets with the smallest and largest average distance error alongside real patient results with visible artefacts such as the movement of hands leading to gross errors.}
\label{fig:colourmaps}
\end{figure}

With a method-average mean surface error of 0.95\,mm versus 3.4\,mm, all methods perform better on the synthetic patient than on the synthetic phantom, even if methods only use depth tracking (see Fig.\ \ref{fig:surface_error_boxplot}). The textureless and perfectly symmetrical phantom makes motion estimation and registration more challenging. ElasticFusion's rotational error in Fig.\,\ref{fig:synthetic_pose_error} is consistent with the perceived loss of tracking towards the end of the synthetic phantom sequence resulting in geometric distortions as a symptom of poor geometric and photometric variation. The boxplot in Fig.\ \ref{fig:surface_error_boxplot} also shows Lacher et al.'s reconstructions to be up to a magnitude more accurate with mean errors of 0.3\,mm and 1.2\,mm exhibiting little misestimation for both synthetic data sequences. Nonetheless, a smear of the nipple in the surface distance error colourmap for Patient 8 in Fig.\,\ref{fig:colourmaps} is hinting a minor in-plane camera drift. VisualSfM failed to reconstruct or did not show sufficient breast coverage in $33\%$ of data sets including all phantom sequences. This is expected as our data violates all major assumptions made by VisualSfM including Lambertian reflectance (skin specularities), illumination invariance (shading varies for self-rotating patient under static illumination) and reliable texture (homogeneous skin). Moreover, VisualSfM discards the temporal order of images, using feature detection and matching instead of feature tracking. Oversaturation in auto-exposed Kinect v2 RGB also makes feature matching more difficult and causes reconstruction gaps. As VisualSfM does not support masks, 2D pixel edges between skin and background are frequently picked up wrongly as salient features leading to background blending artefacts. The inferiority of the results using the RGB-only reconstruction method, highlights the fundamental importance of depth data for 3D reconstruction for this particular application. 
\begin{figure}[t]
\centering
\includegraphics[width=\columnwidth]{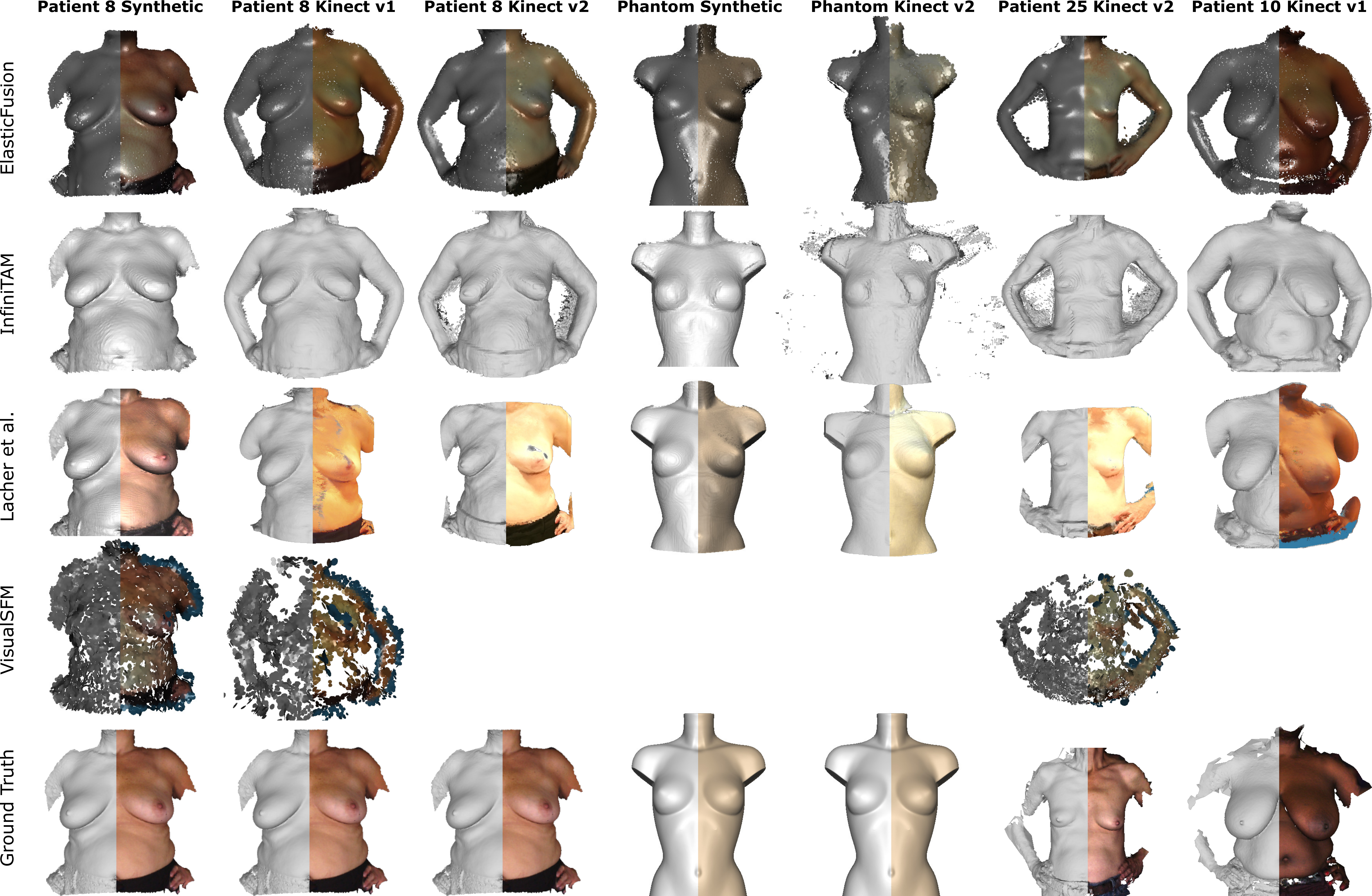}
\caption{Qualitative results in frontal view. Left to right: Synthetic, Kinect v1 and v2 reconstructions for the same patient, phantom reconstructions and two patients of different cup size. Texture, where available, is partly blended onto geometry.}
\label{fig:qualitative_results}
\end{figure}
On top of VisualSfM's consistently higher surface errors averaging to over 10\,mm poor point density is not penalised in the unidirectional surface error metric. As Lacher et\ al.'s solution is tailored to reconstruct a human torso its reconstructed models have less discernible discretization artefacts in comparison to InfiniTAM whose default settings are tuned to larger objects in a fixed-size volume representation. Confining the reconstruction volume to the torso increases resolution and improves registration due to the exclusion of non-rigidly moving parts like the arms. As the motion in the data sequences does not loop, ElasticFusion might have performed below its capabilities being only restricted to rigid tracking. In light of scarce research on clinically acceptable accuracies, InfiniTAM measures a surface error of 3.7$\pm$1.9\,mm over all patient data sets on par with ElasticFusion's 3.9$\pm$1.0\,mm but marginally less accurate than Lacher et\ al.\ with 2.9$\pm$0.9\,mm. Fig. \ref{fig:colourmaps} reveals larger surface errors in the abdominal region in real patient reconstructions. This is likely caused by involuntary non-rigid motions such as breathing, changing hand placement and shoulder torsion. It is also worth of note that Kinect v2, unlike Kinect v1, is affected by flying pixel effects on the boundaries between foreground and background that can result in the reconstruction of small inexistent particles. Despite both being KinectFusion-based methods, Lacher et\ al.\ filter flying pixels and thus produce visibly better reconstructions with Kinect v2 data when compared to InfiniTAM in Fig.\,\ref{fig:qualitative_results}.
No statistically significant difference in accuracy could be established between structured light Kinect v1 and time-of-flight Kinect v2 (p\,=\,0.09). Lacher et\ al.'s results are of superior accuracy compared to all competing methods (p\,$\leq$\,0.01), while VisualSfM performs significantly worse (p\,\textless\,$10^{-9}$).
The design of the clinical acquisition protocol placed its focus on patient safety and least process overhead. In doing so, it introduced two sources of non-rigid deformation. Firstly, patients' self-rotation results in slight articulated motion of body parts and involuntary soft tissue deformity. Secondly, minor posture changes occur between Kinect and \gls{acronym:gt} acquisition. As an indicator of the latter, a residual \gls{acronym:icp} alignment error of 0.9$\pm$0.2\,mm between repeatedly acquired \gls{acronym:gt} scans hints at the extent of non-rigid deformation and sets a realistic lower bound for reported surface-to-surface errors.

\section{Conclusions}

We qualitatively and quantitatively assess four generic 3D reconstruction systems for breast surface modelling from phantom and patient RGBD video in the context of surgical planning and treatment evaluation. Two out of four methods produce submillimeter-accurate results on synthetic and three out of four errors in the order of a few millimeters on clinical data. We believe this to be the first comparison study to focus on a low-cost, infrastructure-less pipeline from acquisition to reconstruction only using consumer market cameras, freely available research software and a standard PC.

\paragraph{\bf Acknowledgements:}

This work was supported by the EPSRC (EP/N013220/1, EP/N022750/1, EP/N027078/1, NS/A000027/1, EP/P012841/1), The Wellcome Trust (WT101957, 201080/Z/16/Z), the EU FP7 VPH-PICTURE (FP7-ICT-2011-9-600948) and Horizon2020 EndoVESPA project (H2020-ICT-2015-688592).

\bibliographystyle{splncs03}
\bibliography{references}

\clearpage

\end{document}